\documentclass{llncs}
\usepackage{makeidx}  
\usepackage{amsmath}
\usepackage{amsfonts}
\usepackage{graphicx}
\usepackage{multirow}
\usepackage[ruled,vlined]{algorithm2e}
\usepackage{epstopdf}

\DeclareMathOperator*{\argmin}{argmin}
\DeclareMathOperator*{\argmax}{argmax}

\begin{document}

\mainmatter              
\title{Alternating direction method of multipliers for regularized  multiclass support vector machines}
\titlerunning{ADMM for regularized MSVM}  
%
\author{Yangyang Xu\inst{1} \and 
Ioannis Akrotirianakis\inst{2} \and Amit Chakraborty\inst{2} }
\authorrunning{Y. Xu, I. Akrotirianakis, A. Chakraborty} 
%
\tocauthor{Yangyang Xu, Ioannis Akrotirianakis, Amit Chakraborty}
\institute{Rice University, Houston, TX, USA\\
\email{yangyang.xu@rice.edu}
\and
Siemens Corporate Technology, Princeton, NJ, USA\\
\email{(ioannis.akrotirianakis, amit.chakraborty)@siemens.com}
}

\maketitle              

\newtheorem{alg}{Algorithm}[section]
\newtheorem{thm}{Theorem}[section]
\newtheorem{cor}{Corollary}[section]
\newtheorem{lem}{Lemma}[section]
\newtheorem{rem}{Remark}[section]
\newcommand{\xbold}{\textbf{x}}
\newcommand{\Xbar}{\bar{X}}
\newcommand{\ybold}{\textbf{y}}
\newcommand{\xboldi}{\textbf{x}_i}
\newcommand{\xboldj}{\textbf{x}_j}
\newcommand{\xibar}{\overline{\xi}}
\newcommand{\alphabar}{\overline{\alpha}}
\newcommand{\alphatild}{\tilde{\alpha}}
\newcommand{\wbold}{\textbf{w}}
\newcommand{\cbold}{\textbf{c}}
\newcommand{\ubold}{\textbf{u}}
\newcommand{\vbold}{\textbf{v}}
\newcommand{\abold}{\textbf{a}}
\newcommand{\ebold}{\textbf{e}}
\newcommand{\zerobold}{\textbf{0}}
\newcommand{\dbold}{\textbf{d}}
\newcommand{\gbold}{\textbf{g}}
\newcommand{\zbold}{\textbf{z}}
\newcommand{\sbold}{\textbf{s}}
\newcommand{\tbold}{\textbf{t}}
\newcommand{\rbold}{\textbf{r}}


\newcommand {\opLip}{\operatorname{Lip}}
\newcommand {\BV}{\operatorname{BV}}
\newcommand {\TV}{\operatorname{TV}}
\newcommand {\Ext}{\operatorname{Ext}}
\newcommand{\tr}{{\mathrm{tr}}}
\newcommand{\diagg}{\mathrm{diag}}
\newcommand{\Diag}{\mathrm{Diag}}
\newcommand{\vc}{\mathrm{vec}}
\newcommand{\prox}{\mathrm{prox}}

\newcommand{\eps}{\epsilon}
\newcommand{\veps}{\varepsilon}
\newcommand{\vphi}{\varphi}
\newcommand{\st}{\text{ subject to }}

\newcommand{\bm}[1]{\boldsymbol{#1}}

\newcommand{\bmpi}{\bm{\Pi}}
\newcommand{\bmxi}{\bm{\Xi}}
\newcommand{\bmtheta}{\bm{\Theta}}
\newcommand{\bmlam}{\bm{\Lambda}}
\newcommand{\bmgam}{\bm{\Gamma}}

\newcommand{\mbc}{\mathbb{C}}
\newcommand{\mbr}{\mathbb{R}}

\newcommand{\mbfa}{\mathbf{A}}
\newcommand{\mbfb}{\mathbf{B}}
\newcommand{\mbfc}{\mathbf{C}}
\newcommand{\mbfd}{\mathbf{D}}
\newcommand{\mbfe}{\mathbf{E}}
\newcommand{\mbfi}{\mathbf{I}}
\newcommand{\mbfp}{\mathbf{P}}
\newcommand{\mbfu}{\mathbf{U}}
\newcommand{\mbfv}{\mathbf{V}}
\newcommand{\mbfw}{\mathbf{W}}
\newcommand{\mbfx}{\mathbf{X}}
\newcommand{\mbfz}{\mathbf{Z}}

\newcommand{\mca}{\mathcal{A}}
\newcommand{\mcc}{\mathcal{C}}
\newcommand{\mcd}{\mathcal{D}}
\newcommand{\mce}{\mathcal{E}}
\newcommand{\mcf}{\mathcal{F}}
\newcommand{\mcg}{\mathcal{G}}
\newcommand{\mci}{\mathcal{I}}
\newcommand{\mcl}{\mathcal{L}}
\newcommand{\mcn}{\mathcal{N}}
\newcommand{\mcp}{\mathcal{P}}
\newcommand{\mcq}{\mathcal{Q}}
\newcommand{\mcs}{\mathcal{S}}
\newcommand{\mct}{\mathcal{T}}
\newcommand{\mcu}{\mathcal{U}}
\newcommand{\mcv}{\mathcal{V}}
\newcommand{\mcx}{\mathcal{X}}

\newcommand{\ma}{{A}}
\newcommand{\opt}{{\mathrm{opt}}}
\newcommand{\conj}{{\mathrm{conj}}}

\def\eop{{\hfill $\spadesuit$}}
\def\sslash{{\backslash\backslash}}
\newcount\refnum\refnum=0
\def\myref{{\global\advance\refnum by 1} {\bf \large Lecture \the \refnum. }}
\newcommand{\bfx}{{\bf x}}
\newcommand{\bfb}{{\bf b}}
\newcommand{\bfc}{{\bf c}}
\newcommand{\bfe}{{\bf e}}
\newcommand{\bfg}{{\bf g}}
\newcommand{\bfy}{{\bf y}}
\newcommand{\bfz}{{\bf z}}
\newcommand{\bfu}{{\bf u}}
\newcommand{\bfv}{{\bf v}}
\newcommand{\bfw}{{\bf w}}

\newcommand{\etal}{{\textit{et al.}}}
\newcommand{\sign}{{\mathrm{sign}}}

\newcommand{\trace}{{\mathrm{tr}}}

\begin{abstract}
The support vector machine (SVM) was originally designed for binary classifications. A lot of effort has been put to generalize the binary SVM to multiclass SVM (MSVM) which are more complex problems. Initially, MSVMs were solved by considering their dual formulations which are quadratic programs and can be solved by standard second-order methods. However, the duals of MSVMs with regularizers are usually more difficult to formulate and computationally very expensive to solve. This paper focuses on several regularized MSVMs and extends the alternating direction method of multiplier (ADMM) to these MSVMs. Using a splitting technique, all considered MSVMs are written as two-block convex programs, for which the ADMM has global convergence guarantees. Numerical experiments on synthetic and real data demonstrate the high efficiency and accuracy of our algorithms.
\end{abstract}
\section{Introduction}\label{sec:motivation}

The linear support vector machine (SVM) \cite{cortes1995support} aims to find a hyperplane to separate a set of data points. It was orginally designed for binary classifications. Motivated by texture classification and gene expression analysis, which usually have a large number of variables but only a few relevant, certain sparsity regularizers such as the $\ell_1$ penalty \cite{bradley1998feature}, need to be included in the SVM model to control the sparsity pattern of the solution and achieve both classification and variable selection. On the other hand, the given data points may belong to more than two classes. To handle the more complex multiclass problems, the binary SVM has been generalized to multicategory classifications \cite{crammer2002algorithmic}. 

The initially proposed multiclass SVM (MSVM) methods construct several binary classifiers, such as ``one-against-one'' \cite{BishopPRMLbook2006}, ``one-against-rest'' \cite{bottou1994comparison} and ``directed acyclic graph SVM'' \cite{platt2000large}. These models are usually solved by considering their dual formulations, which are quadratic programs often with fewer variables and can be efficiently solved by quadratic programming methods. However, these MSVMs may suffer from data imbalance (i.e., some classes have much fewer data points than others) which can result in inaccurate predictions. One alternative is to put all the data points together in one model, which results in the so-called ``all-together'' MSVMs; see \cite{hsu2002comparison} and references therein for the comparison of different MSVMs. The ``all-together'' MSVMs train multi-classifiers by solving one large optimization problem, whose dual formulation is also a quadratic program. In the applications of microarray classification, variable selection is important since most times only a few genes are closely related to certain diseases. Therefore some structure regularizers such as $\ell_1$ penalty \cite{wang20071} and $\ell_\infty$ penalty \cite{zhang2008variable} need to be added to the MSVM models. With the addition of the structure regularizers, the dual problems of the aforementioned MSVMs can be difficult to formulate and hard to solve by standard second-order optimization methods. 

In this paper, we focus on three ``all-together'' regularized MSVMs. 
Specifically, given a set of samples $\{\bfx_i\}_{i=1}^n$ in $p$-dimensional space and each $\bfx_i$ with a label $y_i\in\{1,\cdots,J\}$, we solve the constrained optimization problem
{\small\begin{equation}\label{eq:svms}	
	\underset{\mbfw,\bfb}{\min}\,  \ell_G(\mbfw,\bfb)+\lambda_1\|\mbfw\|_1+\lambda_2 \phi(\mbfw)+\frac{\lambda_3}{2}\|\bfb\|_2^2,
	\text{ s.t. } \mbfw \bfe = \mathbf{0}, \bfe^\top\bfb = 0	
	\end{equation}}
where \vspace{-0.3cm}
{\small $$\ell_G(\mbfw,\bfb)=\frac{1}{n}\sum_{i=1}^n\sum_{j=1}^J I(y_i\neq j)[b_j+\bfw_j^\top\bfx_i+1]_+$$}is generalized hinge loss function;
$I(y_i\neq j)$ equals \emph{one} if $y_i\neq j$ and \emph{zero} otherwise; $[t]_+=\max(0,t)$; $\bfw_j$ denotes the $j$th column of $\mbfw$; $\bfe$ denotes the vector of appropriate size with all \emph{one}s; $\|\mbfw\|_1=\sum_{i,j}|w_{ij}|$; 
$\phi(\mbfw)$ is some regularizer specified below. Usually, the regularizer can promote the structure of the solution and also avoid overfitting problems when the training samples are far less than features. The constraints $\mbfw \bfe = \mathbf{0}, \bfe^\top\bfb = 0$ are imposed to eliminate redundancy in $\mbfw,\bfb$ and are also necessary to make the loss function $\ell_G$ Fisher-consistent \cite{lee2004multicategory}. The solution of \eqref{eq:svms} gives $J$ linear classifiers $f_j(\bfx) = \bfw_j^\top\bfx +b_j,\ j = 1,\cdots,J$. A new coming data point $\bfx$ can be classified by the rule $\text{class}(\bfx)=\argmax_{1\le j \le J}f_j(\bfx)$. 

We consider the following three different forms of $\phi(\mbfw)$:
{\vspace{-0.2cm}\small \begin{subequations}\label{reguw}
		\begin{align}
		\mbox{elastic net: }&\begin{array}{l}\phi(\mbfw)=\frac{1}{2}\|\mbfw\|_F^2,\end{array}\label{eq:elastic}\\
		\mbox{group Lasso: }&\begin{array}{l}\phi(\mbfw)=\sum_{j=1}^p\|\bfw^{j}\|_2,\end{array}\label{eq:splasso}\\
		\mbox{supnorm: }&\begin{array}{l}\phi(\mbfw)=\sum_{j=1}^p\|\bfw^{j}\|_\infty,\end{array}\label{eq:spinf}
		\end{align}
	\end{subequations}}where $\bfw^j$ denotes the $j$th row of $\mbfw$. They fit to data with different structures and can be solved by a \emph{unified} algorithmic framework. Note that we have added the term $\frac{\lambda_3}{2}\|\bfb\|_2^2$ in \eqref{eq:svms}. 
	A positive $\lambda_3$ will make our algorithm more efficient and easier to implement. The extra term usually does not affect the accuracy of classification and variable selection as shown in \cite{hsu2002comparison} for binary classifications. If $\lambda_3$ happens to affect the accuracy, one can choose a tiny $\lambda_3$. 
Model \eqref{eq:svms} includes as special cases the models in \cite{lee2004multicategory} and \cite{zhang2008variable} by letting $\phi$ be the one in \eqref{eq:elastic} and \eqref{eq:spinf} respectively and setting $\lambda_1=\lambda_3=0$. 
To the best of our knowledge, the regularizer \eqref{eq:splasso} has not been considered in MSVM before. It encourages group sparsity of the solution \cite{yuan2006model}, and our experiments will show that \eqref{eq:splasso} can give similar results as those by \eqref{eq:spinf}. Our main contributions are: (i) the development of a unified algorithmic framework based on the ADMM that can solve MSVMs with the three different regularizers defined in \eqref{reguw}; (ii) the proper use of the Woodbury matrix identity \cite{hager1989updating} which can reduce the size of the linear systems arising during the solution of \eqref{eq:svms}; (iii) computational experiments on a variety of datasets that practically demonstrate that our algorithms can solve large-scale multiclass classification problems much faster than state-of-the-art second order methods.

	We use $\bfe$ and $\mbfe$ to denote a vector and a matrix with all \emph{one}s, respectively. $\mbfi$ is used for an identity matrix. Their sizes are clear from the context.
	
	The rest of the paper is organized as follows. 
	Section \ref{sec:application} gives our algorithm for solving \eqref{eq:svms}. 
Numerical results are given in section \ref{sec:numerical} on both synthetic and real data. Finally, section \ref{sec:conclusion} concludes this paper.

	\section{Algorithms}\label{sec:application}
	In this section we extend ADMM into the general optimization problems described by \eqref{eq:svms}. Due to lack of space we refer the reader to 
	\cite{boydADMMbook} for details of ADMM. 
	We first consider \eqref{eq:svms} with $\phi(\mbfw)$ defined in \eqref{eq:elastic} and then solve it with $\phi(\mbfw)$ in \eqref{eq:splasso} and \eqref{eq:spinf} in a unified form. The parameter $\lambda_3$ is always assumed positive.	
	One can also transform the MSVMs to quadratic or second-order cone programs and use standard second-order methods to solve them. Nevertheless, these methods are computationally intensive for large-scale problems. As shown in section \ref{sec:numerical}, ADMM is, in general, much faster than standard second-order methods. 
	\subsection{ADMM for solving \eqref{eq:svms} with $\phi$ defined by \eqref{eq:elastic}}\label{sec:solveelastic}
	Introduce auxiliary variables $\mbfa=\mbfx^\top\mbfw+\bfe\bfb^\top+\mbfe$ and $\mbfu=\mbfw$, where $\mbfx=[\bfx_1,\cdots,\bfx_n]\in\mbr^{p\times n}$. 
	Using the above auxiliary variables we can equivalently write \eqref{eq:svms} with $\phi(\mbfw)$ defined in \eqref{eq:elastic} as follows
	{\small \begin{equation}\label{eq:elastic2}
		\begin{array}{cl}
		{\min}&\frac{1}{n}\underset{i,j}{\sum}c_{ij}[a_{ij}]_++\lambda_1\|\mbfu\|_1+\frac{\lambda_2}{2}\|\mbfw\|_F^2+\frac{\lambda_3}{2}\|\bfb\|_2^2\\[0.1cm]
		\text{s.t.} & \mbfa=\mbfx^\top\mbfw+\bfe\bfb^\top+\mbfe,\ \mbfu=\mbfw, \mbfw\bfe=\mathbf{0},\ \bfe^\top\bfb=0.\vspace{-0.2cm}
		\end{array}
		\end{equation}}
	The augmented Lagrangian\footnote{We do not include the constraints $\mbfw\bfe=\mathbf{0},\ \bfe^\top\bfb=0$ in the augmented Lagrangian, but instead we include them in $(\mbfw,\bfb)$-subproblem; see the update \eqref{elastic_wb}.} of \eqref{eq:elastic2} is
	{\small\begin{equation}\label{lag:elastic}
		\begin{array}{rl}
		\mcl_1(\mbfw,\bfb,\mbfa,\mbfu,\bmpi,\bmlam)
		=&\frac{1}{n}\underset{i,j}{\sum}c_{ij}[a_{ij}]_++\lambda_1\|\mbfu\|_1+\frac{\lambda_2}{2}\|\mbfw\|_F^2+\frac{\lambda_3}{2}\|\bfb\|_2^2+\langle\bmlam, \mbfw-\mbfu\rangle\\[0.1cm]
		&+\frac{\mu}{2}\|\mbfw-\mbfu\|_F^2
		+\langle\bmpi, \mbfx^\top\mbfw+\bfe\bfb^\top-\mbfa+\mbfe\rangle\\[0.1cm]
		&+\frac{\alpha}{2}\|\mbfx^\top\mbfw+\bfe\bfb^\top-\mbfa+\mbfe\|_F^2,
		\end{array}
		\end{equation}}where $\bmpi,\bmlam$ are Lagrange multipliers and $\alpha,\mu>0$ are penalty parameters. The ADMM approach for \eqref{eq:elastic2} can be derived by minimizing $\mcl_1$ alternatively with respect to $(\mbfw,\bfb)$ and $(\mbfa,\mbfu)$ and updating the multipliers $\bmpi,\bmlam$, namely, at iteration $k$,
	{\small \begin{subequations}\label{admm:elastic}
			\begin{eqnarray}
			&&\big(\mbfw^{(k+1)},\bfb^{(k+1)}\big)=\underset{(\mbfw,\bfb)\in\mcd}{\argmin}\mcl_1\big(\mbfw,\bfb,\mbfa^{(k)},\mbfu^{(k)},\bmpi^{(k)},\bmlam^{(k)}\big),\label{elastic_wb}\\
			&&\big(\mbfa^{(k+1)},\mbfu^{(k+1)}\big)
			=\underset{\mbfa,\mbfu}{\argmin}\ \mcl_1\big(\mbfw^{(k+1)},\bfb^{(k+1)},\mbfa,\mbfu,\bmpi^{(k)},\bmlam^{(k)}\big),\label{elastic_au}\\			&&\bmpi^{(k+1)}=\bmpi^{(k)}+\alpha\big(\mbfx^\top\mbfw^{(k+1)}
			+\bfe(\bfb^{(k+1)})^\top-\mbfa^{(k+1)}+\mbfe\big),\label{elastic_pi}\\
			&&\bmlam^{(k+1)}
			=\bmlam^{(k)}+\mu\big(\mbfw^{(k+1)}-\mbfu^{(k+1)}\big),\label{elastic_lam}
			\end{eqnarray}
		\end{subequations}}where $\mcd=\{(\mbfw,\bfb):\ \mbfw\bfe=\mathbf{0},\ \bfe^\top\bfb=0\}$. The updates \eqref{elastic_pi} and \eqref{elastic_lam} are simple. We next discuss how to solve \eqref{elastic_wb} and \eqref{elastic_au}.
		
		\subsubsection{Solution of \eqref{elastic_wb}:}
		Define $\mbfp=[\mbfi; -\bfe^\top]\in \mbr^{J\times(J-1)}$. Let $\hat{\mbfw}$ be the submatrix consisting of the first $J-1$ columns of $\mbfw$ and $\hat{\bfb}$ be the subvector consisting of the first $J-1$ components of $\bfb$. Then it is easy to verify that $\mbfw=\hat{\mbfw}\mbfp^\top, \bfb=\mbfp\hat{\bfb}$ and problem \eqref{elastic_wb} is equivalent to the unconstrained optimization problem
		{\small \begin{equation}\label{temp1}
			\begin{array}{ll}\underset{\hat{\mbfw},\hat{\bfb}}{\min}&\frac{\lambda_2}{2}\|\hat{\mbfw}\mbfp^\top\|_F^2+\langle\bmlam^{(k)}, \hat{\mbfw}\mbfp^\top\rangle+\frac{\lambda_3}{2}\|\hat{\bfb}^\top\mbfp^\top\|_2^2+\frac{\mu}{2}\|\hat{\mbfw}\mbfp^\top-\mbfu^{(k)}\|_F^2\\[0.1cm]
			&+\langle\bmpi^{(k)}, \mbfx^\top\hat{\mbfw}\mbfp^\top+\bfe\hat{\bfb}^\top\mbfp^\top\rangle
+\frac{\alpha}{2}\|\mbfx^\top\hat{\mbfw}\mbfp^\top+\bfe\hat{\bfb}^\top\mbfp^\top-\mbfa^{(k)}+\mbfe\|_F^2.
			\end{array}
			\end{equation}}The first-order optimality condition of \eqref{temp1} is the linear system
			{\small \begin{equation}\label{lineq}
				\begin{bmatrix}
				\alpha\mbfx\mbfx^\top+(\lambda_2+\mu)\mbfi & \alpha\mbfx\bfe\\
				\alpha\bfe^\top\mbfx^\top & n\alpha+\lambda_3
				\end{bmatrix}
				\begin{bmatrix}
				\hat{\mbfw}\\
				\hat{\bfb}^\top
				\end{bmatrix}=\begin{bmatrix}\left(\mbfx
				\bmtheta-\bmlam^{(k)}+\mu\mbfu^{(k)}\right)
				\mbfp(\mbfp^\top\mbfp)^{-1}\\
				\bfe^\top\left(\alpha\mbfa^{(k)}-\bmpi^{(k)}-\alpha\mbfe\right)\mbfp(\mbfp^\top\mbfp)^{-1}
				\end{bmatrix},
				\end{equation}}where $\bmtheta=\alpha\mbfa^{(k)}-\bmpi^{(k)}-\alpha\mbfe$. The size of \eqref{lineq} is $(p+1)\times(p+1)$ and when $p$ is small, we can afford to directly solve it.
However,	if $p$ is large, even the iterative method for linear system (e.g., preconditioned conjugate gradient) can be very expensive. In the case of ``large $p$, small $n$'', we can employ the \emph{Woodbury matrix identity} (e.g., \cite{hager1989updating}) to efficiently solve \eqref{lineq}. In particular, let $\mbfd=\text{block\_diag}((\lambda_2+\mu)\mbfi, \lambda_3)$ and $\mbfz=[\mbfx; \bfe^\top]$. 
Then the coefficient matrix of \eqref{lineq} is $\mbfd+\alpha\mbfz\mbfz^\top$, and by the \emph{Woodbury matrix identity}, we have $\mbfp(\mbfp^\top\mbfp)^{-1}=[\mbfi;\mathbf{0}]-\frac{1}{J}\mbfe$ and 
		{\small\begin{equation*}\label{woodbury}
			\begin{array}{rl}(\mbfd+\alpha\mbfz\mbfz^\top)^{-1}
			=\mbfd^{-1}-\alpha\mbfd^{-1}\mbfz
	(\mbfi+\alpha\mbfz^\top\mbfd^{-1}\mbfz)^{-1}\mbfz^\top\mbfd^{-1}
			\end{array}.
			\vspace{-0.2cm}\end{equation*}}
			
			Note $\mbfd$ is diagonal, and thus $\mbfd^{-1}$ is simple to compute. $\mbfi+\alpha\mbfz^\top\mbfd^{-1}\mbfz$ is $n\times n$ and positive definite. 							
Hence, as $n\ll p$, \eqref{lineq} can be solved by solving a much smaller linear system and doing several matrix-matrix multiplications. In case of large $n$ and $p$, one can perform a proximal gradient step to update $\mbfw$ and $\bfb$, which results in a proximal-ADMM \cite{deng2012}. To the best of our knowledge, this is the first time that the Woodbury matrix identity is used to substantially reduce\footnote{For the case of $n\ll p$, we found that using the Woodbury matrix identity can be about 100 times faster than preconditioned conjugate gradient (pcg) with moderate tolerance $10^{-6}$ for the solving the linear system \eqref{lineq}.} the computational work and allow ADMM to efficiently solve large-scale multiclass SVMs. Solve \eqref{lineq} by multiplying $(\mbfd+\alpha\mbfz\mbfz^\top)^{-1}$ to both sides. Letting $\mbfw^{(k+1)}=\hat{\mbfw}\mbfp^\top$ and $\bfb^{(k+1)}=\mbfp\hat{\bfb}$ gives the solution of \eqref{elastic_wb}.
		
		\subsubsection{Solution of \eqref{elastic_au}:}
		
		Note that $\mbfa$ and $\mbfu$ are independent of each other as $\mbfw$ and $\bfb$ are fixed. Hence we can separately update $\mbfa$ and $\mbfu$ by
		{\vspace{-0.1cm}\small\begin{align*}
			&\mbfa^{(k+1)}=\underset{\mbfa}{\argmin}\frac{1}{n}\sum_{i,j}c_{ij}[a_{ij}]_++\frac{\alpha}{2}\big\|\mbfx^\top\mbfw^{(k+1)}+\bfe(\bfb^{(k+1)})^\top+\frac{1}{\alpha}\bmpi^{(k)}+\mbfe-\mbfa\big\|_F^2\\[-0.1cm]
			&\mbfu^{(k+1)}=\underset{\mbfu}{\argmin}\lambda_1\|\mbfu\|_1+\frac{\mu}{2}\big\|\mbfw^{(k+1)}+\frac{1}{\mu}\bmlam^{(k)}-\mbfu\big\|_F^2.
			\end{align*}}Both the above problems are separable and have closed form solutions
		{\vspace{-0.2cm}\small
			\begin{align}
			&a_{ij}^{(k+1)}
			=\mct_{\frac{c_{ij}}{n\alpha}}\left(\left(\mbfx^\top\mbfw^{(k+1)}+
			\bfe(\bfb^{(k+1)})^\top+\frac{1}{\alpha}\bmpi^{(k)}+\mbfe\right)_{ij}\right),\,\forall i,j,\label{sol:elastic_a}\\
			&u_{ij}^{(k+1)}=\mcs_{\frac{\lambda_1}{\mu}}\left(\left(\mbfw^{(k+1)}+\frac{1}{\mu}\bmlam^{(k)}\right)_{ij}\right),\ \forall i, j,\label{sol:elastic_u}
			\end{align}}where {\vspace{-0.2cm}\small $$\mct_\nu(\delta)=\left\{
			\begin{array}{ll}
			\delta-\nu, & \delta>\nu,\\
			0, &0\le\delta\le\nu,\\
			\delta,& \delta<0,
			\end{array}\right.$$}and $\mcs_\nu(\delta)=\sign(\delta)\max(0,|\delta|-\nu)$.	
		Putting the above discussions together, we have Algorithm \ref{alg:elastic} for solving  \eqref{eq:svms} with $\phi$ defined in \eqref{eq:elastic}. 
		{\small \vspace{-0.5cm}
	\begin{algorithm}\caption{ADMM for \eqref{eq:svms} with $\phi(\mbfw)$ in \eqref{eq:elastic}}\label{alg:elastic}
		\DontPrintSemicolon
{\bf Input:} $n$ sample-label pairs $\{(\bfx_i,y_i)\}_{i=1}^n$.\;
				{\bf Choose:} $\alpha,\mu>0$ and $(\mbfw_0,\bfb_0,\mbfa_0,\mbfu_0,\bmpi_0,\bmlam_0),k=0$.\;
				\While{not converge}{
				Solve \eqref{lineq}; let $\mbfw^{(k+1)}=\hat{\mbfw}\mbfp^\top$ and $\bfb^{(k+1)}=\mbfp\hat{\bfb}$;\;
				Update $\mbfa^{(k+1)}$ and $\mbfu^{(k+1)}$ by \eqref{sol:elastic_a} and \eqref{sol:elastic_u};\;
				Update $\bmpi^{(k+1)}$ and $\bmlam^{(k+1)}$ by \eqref{elastic_pi} and \eqref{elastic_lam};\;
				Let $k=k+1$
				}
	\end{algorithm} } \vspace{-0.75cm}

\subsection{ADMM for solving \eqref{eq:svms} with $\phi$ defined by \eqref{eq:splasso} and \eqref{eq:spinf}}
Firstly, we write \eqref{eq:svms} with $\phi(\mbfw)$ defined in \eqref{eq:splasso} and \eqref{eq:spinf} in the unified form of
{\vspace{-0.1cm}\small\begin{equation}\label{eq:spqnorm}
	\underset{\mbfw,\bfb}{\min}\, \ell_G(\mbfw,\bfb)+\lambda_1\|\mbfw\|_1+\overset{p}{\underset{j=1}{\sum}}{\lambda_2}\|\bfw^{j}\|_q+\frac{\lambda_3}{2}\|\bfb\|_2^2,\ 
	\text{s.t.} \mbfw \bfe = \mathbf{0}, \bfe^\top\bfb = 0,
	\end{equation}}where $q=2$ for \eqref{eq:splasso} and $q=\infty$ for \eqref{eq:spinf}. Introducing auxiliary variables $\mbfa=\mbfx^\top\mbfw+\bfe\bfb^\top+\mbfe$, $\mbfu=\mbfw$, and $\mbfv=\mbfw$, we can write \eqref{eq:spqnorm} equivalently to
{\vspace{-0.2cm}\small\begin{equation}\label{eq:spqnorm2}
	\begin{array}{cl}
	{\min}&\frac{1}{n}\underset{i,j}{\sum}c_{ij}[a_{ij}]_++\lambda_1\|\mbfu\|_1+\overset{p}{\underset{j=1}{\sum}}\lambda_2\|\bfv^j\|_q+\frac{\lambda_3}{2}\|\bfb\|^2\\[0.1cm]
	\text{s.t.} &\mbfa=\mbfx^\top\mbfw+\bfe\bfb^\top+\mbfe,\ \mbfu=\mbfw,\ \mbfv=\mbfw,\ \mbfw\bfe=\mathbf{0},\ \bfe^\top\bfb=0.
	\vspace{-0.2cm}\end{array}
	\end{equation}}
The augmented Lagrangian of \eqref{eq:spqnorm2} is
{\vspace{-0.2cm}\small\begin{align}
	\mcl_2(\mbfw,\bfb,\mbfa,\mbfu,\mbfv,\bmpi,\bmlam,\bmgam)
	=&\frac{1}{n}\underset{i,j}{\sum}c_{ij}[a_{ij}]_++\lambda_1\|\mbfu\|_1+\overset{p}{\underset{j=1}{\sum}}\lambda_2\|\bfv^j\|_q+\frac{\lambda_3}{2}\|\bfb\|_2^2\nonumber\\[-0.1cm]
	&\hspace{-2.5cm}+\langle\bmpi, \mbfx^\top\mbfw+\bfe\bfb^\top-\mbfa+\mbfe\rangle+\frac{\alpha}{2}\|\mbfx^\top\mbfw+\bfe\bfb^\top-\mbfa+\mbfe\|_F^2\nonumber\\[-0.1cm]
	&\hspace{-2.5cm}+\langle\bmlam, \mbfw-\mbfu\rangle+\frac{\mu}{2}\|\mbfw-\mbfu\|_F^2+\langle\bmgam, \mbfw-\mbfv\rangle+\frac{\nu}{2}\|\mbfw-\mbfv\|_F^2,\label{eq:spqnorm2_Lagrangian}
	\end{align}}where $\bmpi,\bmlam,\bmgam$ are Lagrange multipliers and $\alpha,\mu,\nu>0$ are penalty parameters.
The ADMM updates for \eqref{eq:spqnorm2} can be derived as 
{\vspace{-0.2cm}\small\begin{subequations}\label{admm:spqnorm}
		\begin{align}
		\big(\mbfw^{(k+1)},\bfb^{(k+1)}\big)=&\underset{(\mbfw,\bfb)\in\mcd}{\argmin}\mcl_2\big(\mbfw,\bfb,\mbfa^{(k)},\mbfu^{(k)},\mbfv^{(k)},\bmpi^{(k)},\bmlam^{(k)},\bmgam^{(k)}\big)\label{spqnorm_wb}\\
		\big(\mbfa^{(k+1)},\mbfu^{(k+1)},\mbfv^{(k+1)}\big)=&\underset{\mbfa,\mbfu,\mbfv}{\argmin}\,\mcl_2\big(\mbfw^{(k+1)},\bfb^{(k+1)},\mbfa,\mbfu,\mbfv,\bmpi^{(k)},\bmlam^{(k)},\bmgam^{(k)}\big)\label{spqnorm_auv}\\	\bmpi^{(k+1)}=&\bmpi^{(k)}+\alpha\big(\mbfx^\top\mbfw^{(k+1)}		+\bfe(\bfb^{(k+1)})^\top-\mbfa^{(k+1)}+\mbfe\big)\label{spqnorm_pi}\\
\bmlam^{(k+1)}=&\bmlam^{(k)}+\mu\big(\mbfw^{(k+1)}-\mbfu^{(k+1)}\big),\label{spqnorm_lam}\\
\bmgam^{(k+1)}=&\bmgam^{(k)}+\nu\big(\mbfw^{(k+1)}-\mbfv^{(k+1)}\big).\label{spqnorm_gam}
		\end{align}
	\end{subequations}}The subproblem \eqref{spqnorm_wb} can be solved in a similar way as discussed in section \ref{sec:solveelastic}. Specifically, first obtain $(\hat{\mbfw},\hat{\bfb})$ by solving 
		{\vspace{-0.1cm}\small
			\begin{equation}\label{lineq2}
			\begin{bmatrix}
			\alpha\mbfx\mbfx^\top+(\nu+\mu)\mbfi & \alpha\mbfx\bfe\\
			\alpha\bfe^\top\mbfx^\top & n\alpha+\lambda_3
			\end{bmatrix}
			\begin{bmatrix}
			\hat{\mbfw}\\
			\hat{\bfb}^\top
			\end{bmatrix}= 
			\begin{bmatrix}
			\begin{array}{l}\big(\mbfx
			\bmxi-\bmlam^{(k)}-\bmgam^{(k)}+\mu\mbfu^{(k)}+\nu\mbfv^{(k)}\big)
			\mbfp(\mbfp^\top\mbfp)^{-1}
			\end{array}
			\\
			\bfe^\top\left(\alpha\mbfa^{(k)}-\bmpi^{(k)}-\alpha\mbfe\right)\mbfp(\mbfp^\top\mbfp)^{-1}
			\end{bmatrix},
			\end{equation}
		}where $\bmxi=\alpha\mbfa^{(k)}-\bmpi^{(k)}-\alpha\mbfe$ and then let $\mbfw^{(k+1)}=\hat{\mbfw}\mbfp^\top,~\bfb^{(k+1)}=\mbfp\hat{\bfb}$.
	To solve \eqref{spqnorm_auv} note that $\mbfa, \mbfu$ and $\mbfv$ are independent of each other and can be updated separately. The update of $\mbfa$ and $\mbfu$ is similar to that described in section \ref{sec:solveelastic}. We next discuss how to update $\mbfv$ by solving the problem
	{\vspace{-0.2cm}\small\begin{equation}\label{spqnorm_v}
		\mbfv^{(k+1)}=\argmin_\mbfv \sum_{j=1}^p\lambda_2\|\bfv^j\|_q+\frac{\nu}{2}\big\|\mbfw^{(k+1)}+\frac{1}{\nu}\bmgam^{(k)}-\mbfv\big\|_F^2
		\end{equation}}Let $\mbfz=\mbfw^{(k+1)}+\frac{1}{\nu}\bmgam^{(k)}$. According to \cite{yuan2006model}, the solution of \eqref{spqnorm_v} for $q=2$ is
	{\vspace{-0.1cm}\small\begin{equation}\label{sol:spqnorm_v}
		\left(\bfv^{(k+1)}\right)^j=\left\{
		\begin{array}{cl}\mathbf{0},& \|\bfz^j\|_2\le\frac{\lambda_2}{\nu}\\[0.2cm]
		\frac{ \|\bfz^j\|_2-\lambda_2/\nu}{\|\bfz^j\|_2}\bfz^j,& \text{ otherwise}
		\end{array}\right.,\forall j.
		\end{equation}}For $q=\infty$, the solution of \eqref{spqnorm_v} can be computed via Algorithm \ref{alg:spqnorm_v} (see  \cite{chen2009accelerated} for details). Putting the above discussions together, we have Algorithm \ref{alg:splasso_inf} for solving \eqref{eq:svms} with $\phi(\mbfw)$ given by \eqref{eq:splasso} and \eqref{eq:spinf}.

		{\small \vspace{-0.3cm}
	\begin{algorithm}\caption{Algorithm for solving \eqref{spqnorm_v} when $q=\infty$}
		\label{alg:spqnorm_v}
		\DontPrintSemicolon
				{\bf Let} $\tilde{\lambda}=\frac{\lambda_2}{\nu}$ and $\mbfz=\mbfw^{(k+1)}+\frac{1}{\nu}\bmgam^{(k)}$.\;
				\For{$j=1,\cdots,p$}{
				Let $\bfv=\bfz^j$;\;
				\If{$\|\bfv\|_1\le\tilde{\lambda}$}{
				Set $\left(\bfv^{(k+1)}\right)^j=\mathbf{0}$.
				}
				\Else{
				 Let $\bfu$ be the sorted absolute value vector of $\bfv$: $u_1\ge u_2\ge\cdots\ge u_J$;\;
				Find $\hat{r}=\max\left\{r:\tilde{\lambda}-\sum_{t=1}^r(u_t-u_r)>0\right\}$\;
				Let $v_{ji}^{(k+1)}=\sign(v_i)\min\left(|v_i|,(\sum_{t=1}^{\hat{r}}u_t-\tilde{\lambda})/\hat{r}\right),\forall i$.\;
				}
				}
		\end{algorithm}}\vspace{-0.0cm}

	\subsection{Convergence results}
	Let us denote the $k$th iteration of the objectives of \eqref{eq:elastic2} and \eqref{eq:spqnorm2} as
	{\small \vspace{-0.1cm} \begin{equation}\label{notationF}
		\begin{array}{l}F_1^{(k)}=F_1\big(\mbfw^{(k)},\bfb^{(k)},\mbfa^{(k)},\mbfu^{(k)}\big),\
		F_2^{(k)}=F_2\big(\mbfw^{(k)},\bfb^{(k)},\mbfa^{(k)},\mbfu^{(k)},\mbfv^{(k)}\big),\end{array}\end{equation}} and define
		{\vspace{-0.2cm}\small $$\begin{array}{l}\mbfz_1^{(k)}=\big(\mbfw^{(k)},\bfb^{(k)},\mbfa^{(k)},\mbfu^{(k)},\bmpi^{(k)},\bmlam^{(k)}\big),\\[0.1cm]
			\mbfz_2^{(k)}=\big(\mbfw^{(k)},\bfb^{(k)},\mbfa^{(k)},\mbfu^{(k)},\mbfv^{(k)},\bmpi^{(k)},\bmlam^{(k)},\bmgam^{(k)}\big).\end{array}$$}
	\begin{theorem}\label{thm:svms}
		Let $\{\mbfz_1^{(k)}\}$ and $\{\mbfz_2^{(k)}\}$ be the sequences generated by \eqref{admm:elastic} and \eqref{admm:spqnorm}, respectively. Then $F_1^{(k)}\to F_1^*$, $F_2^{(k)}\to F_2^*$, and $\|\mbfx^\top\mbfw^{(k)}+\bfe(\bfb^{(k)})^\top+\mbfe-\mbfa^{(k)}\|_F,\|\mbfw^{(k)}-\mbfu^{(k)}\|_F,\|\mbfw^{(k)}-\mbfv^{(k)}\|_F$ all converge to \emph{zero}, where $F_1^*$ and $F_2^*$ are the optimal objective values of \eqref{eq:elastic2} and \eqref{eq:spqnorm2}, respectively. In addition, if $\lambda_2>0, \lambda_3>0$ in \eqref{eq:elastic2}, then $\mbfz_1^{(k)}$ converges linearly.
	\end{theorem}\vspace{-0.0cm}
	
	The proof is based on \cite{glowinski2008numerical,deng2012} and due to the lack of space we omit it.

%

		\section{Numerical results}\label{sec:numerical}
		We now test the three different regularizers in \eqref{reguw} on two sets of synthetic data and two sets of real data. As shown in \cite{wang20071} the $L_1$ regularized MSVM works better than the standard ``one-against-rest'' MSVM in both classification and variable selection. Hence, we choose to only compare the three regularized MSVMs. The ADMM algorithms discussed in section \ref{sec:application} are used to solve the three models. Until the preparation of this paper, we did not find much work on designing specific algorithms to solve the regularized MSVMs except \cite{wang20071} which uses a path-following algorithm to solve the $L_1$ MSVM.  To illustrate the efficiency of ADMM, we compare it with Sedumi \cite{sturm1999using} which is a second-order method. We call Sedumi in the CVX environment \cite{grant2008cvx}.
		
				{\small \vspace{-0.1cm}
					\begin{algorithm}\caption{ADMM for \eqref{eq:svms} with $\phi(\mbfw)$ in \eqref{eq:splasso} and \eqref{eq:spinf}}\label{alg:splasso_inf}
						\DontPrintSemicolon
						{\bf Input:} $n$ sample-label pairs $\{(\bfx_i,y_i)\}_{i=1}^n$.\;
						{\bf Choose:} $\alpha,\mu,\nu>0$, set $k=0$ and initialize $(\mbfw_0,\bfb_0,\mbfa_0,\mbfu_0,\mbfv_0,\bmpi_0,\bmlam_0,\bmgam_0)$.\;
						\While{not converge}{
							Solve \eqref{lineq2}; let $\mbfw^{(k+1)}=\hat{\mbfw}\mbfp^\top$ and $\bfb^{(k+1)}=\mbfp\hat{\bfb}$;\;
							Update $\mbfa^{(k+1)}$ and $\mbfu^{(k+1)}$ by \eqref{sol:elastic_a} and \eqref{sol:elastic_u};\;
							Update $\mbfv^{(k+1)}$ by \eqref{sol:spqnorm_v} if $q=2$ and by Algorithm \ref{alg:spqnorm_v} if $q=\infty$;\;
							Update $\bmpi^{(k+1)}$, $\bmlam^{(k+1)}$ and $\bmgam^{(k+1)}$ by \eqref{spqnorm_pi}, \eqref{spqnorm_lam} and \eqref{spqnorm_gam};
						}
					\end{algorithm}} \vspace{-0.1cm}


\subsection{Implementation details}
All our code was written in MATLAB, except the part of Algorithm \ref{alg:spqnorm_v} which was written in C with MATLAB interface. We used $\lambda_3=1$ for all three models. In our experiments, we found that the penalty parameters were very important for the speed of ADMM.
By running a large set of random tests, we chose $\alpha = \frac{50J}{n}, \mu=\sqrt{pJ}$ in \eqref{lag:elastic} and $\alpha = \frac{50J}{n}, \mu=\nu=\sqrt{pJ}$ in \eqref{eq:spqnorm2_Lagrangian}. Origins were used as the starting points. As did in \cite{ye2011efficient}, we terminated ADMM for \eqref{eq:elastic2}, that is, \eqref{eq:svms} with $\phi(\mbfw)$ in \eqref{eq:elastic}, if
{\small $$\vspace{-0.2cm}\max\left\{\begin{array}{l}\frac{\big|F_1^{(k+1)}-F_1^{(k)}\big|}{1+F_1^{(k)}},\  \frac{\big\|\mbfw^{(k)}-\mbfu^{(k)}\big\|_F}{\sqrt{pJ}},
\frac{\big\|\mbfx^\top\mbfw^{(k)}+\bfe(\bfb^{(k)})^\top+\mbfe-\mbfa^{(k)}\big\|_F}{\sqrt{nJ}}\end{array}\right\}\le 10^{-5},$$}and ADMM for \eqref{eq:spqnorm2}, that is, \eqref{eq:svms} with $\phi(\mbfw)$ in \eqref{eq:splasso} and \eqref{eq:spinf}, if
{\small $$\vspace{-0.2cm}\max\left\{\begin{array}{l}\frac{\big|F_2^{(k+1)}-F_2^{(k)}\big|}{1+F_2^{(k)}}, \frac{\big\|\mbfx^\top\mbfw^{(k)}+\bfe(\bfb^{(k)})^\top+\mbfe-\mbfa^{(k)}\big\|_F}{\sqrt{nJ}}, \frac{\big\|\mbfw^{(k)}-\mbfu^{(k)}\big\|_F}{\sqrt{pJ}},\ \frac{\big\|\mbfw^{(k)}-\mbfv^{(k)}\big\|_F}{\sqrt{pJ}}\end{array}\right\}\le 10^{-5}.$$}In addition, we set a maximum number of iterations $maxit=5000$ for ADMM. Default settings were used for Sedumi.  
All the tests were performed on a PC with an i5-2500 CPU and 3-GB RAM and running 32-bit Windows XP. 

\begin{table}\caption{Results of different models solved by ADMM and Sedumi on a five-class example with synthetic data. The numbers in parentheses are standard errors.}
	\label{table:syn1}
	{\small
		\begin{center}
			\scalebox{0.8}{\begin{tabular}{|c|ccccc|ccccc|}\hline
					\multirow{2}{*}{Models}& \multicolumn{5}{|c|}{ADMM} & \multicolumn{5}{|c|}{Sedumi}\\\cline{2-11}
					& Accuracy & time & CZ & IZ & NR & Accuracy & time & CZ & IZ & NR\\\hline
					elastic net & 0.597(0.012) & 0.184 & 39.98 & 0.92 & 2.01 & 0.592(0.013) & 0.378 & 39.94 & 1.05 & 2.03 \\
					group Lasso & 0.605(0.006) & 0.235 & 34.94 & 0.00 & 3.14 & 0.599(0.008) & 2.250 & 33.85 & 0.02 & 3.25 \\
					supnorm & 0.606(0.006) & 0.183 & 39.84 & 0.56 & 2.08 & 0.601(0.008) & 0.638 & 39.49 & 0.61 & 2.21\\\hline 
				\end{tabular}}
			\end{center}}\vspace{-1cm}
	\end{table}									
									
									\subsection{Synthetic data}
									The first test is a five-class example with each sample $\bfx$ in a 10-dimensional space. The data was generated in the following way: for each class $j$, the first two components $(x_1, x_2)$ were generated from the mixture Gaussian distribution $\mcn(\bm{\mu}_j, 2\mbfi)$ where for $j = 1,\cdots,5,$
									\[
									\bm{\mu}_j=2[\cos\left((2j-1)\pi/5\right), \sin\left((2j-1)\pi/5\right)],
									\]
									and the remaining eight components were independently generated from standard Gaussian distribution. This kind of data was also tested in \cite{wang20071,zhang2008variable}. We first chose best parameters for each model by generating $n=200$ samples for training and another $n=200$ samples for tuning parameters. 
For elastic net, we fixed $\lambda_2=1$ since it is not sensitive and then searched the best $\lambda_1$ over $\mcc=\{0,0.001,0.01:0.01:0.1,0.15,0.20,0.25,0.30\}$. The parameters $\lambda_1$ and $\lambda_2$ for group Lasso and supnorm were selected via a grid search over $\mcc\times\mcc$. With the tuned parameters, we compared ADMM and Sedumi on $n=200$ randomly generated training samples and $n'=50,000$ random testing samples, and the whole process was independently repeated 100 times. The performance of the compared models and algorithms were measured by accuracy (i.e., $\frac{\text{number of correctly predicted}}{\text{total number}}$), running time (sec), the number of correct zeros (CZ), the number of incorrect zeros (IZ) and the number of non-zero rows (NR). We counted CZ, IZ and NR from the truncated solution $\mbfw^t$, which was obtained from the output solution $\mbfw$ such that $w^t_{ij}=0$ if $|w_{ij}|\le 10^{-3}\max_{i,j}|w_{ij}|$ and $w^t_{ij}=w_{ij}$ otherwise. The average results are shown in Table \ref{table:syn1}, from which we can see that ADMM produces similar results as those by Sedumi within less time. Elastic net makes slightly lower prediction accuracy than that by the other two models.
		\begin{table}\caption{Results of different models solved by ADMM and Sedumi on a four-class example with synthetic data. The numbers in the parentheses are corresponding standard errors.}
		\label{table:syn2}
		{\small
			\begin{center}
				\scalebox{0.8}{\begin{tabular}{|c|ccccccc||ccccccc|}\hline
						\multirow{3}{*}{Models}& \multicolumn{7}{|c||}{ADMM} & \multicolumn{7}{|c|}{Sedumi}\\\cline{2-15}
						&Accuracy & time  & IZ & NZ1 & NZ2 & NZ3 & NZ4 & Accuracy & time  & IZ & NZ1 & NZ2 & NZ3 & NZ4\\\cline{2-15}
						&\multicolumn{14}{|c|}{Correlation $\rho=0$}\\\hline
						elastic net & 0.977(0.006) & 0.27 & 13.8 & 37.6 & 36.9 & 36.8 & 37.0& 0.950(0.013) & 3.75 & 11.0 & 40.2 & 40.0 & 39.5 & 40.4\\
						group Lasso & 0.931(0.020) & 0.46 & 30.4 & 33.7 & 33.4 & 33.2 & 33.2  & 0.857(0.022) & 12.13 & 40.5 & 31.8 & 31.6 & 31.8 & 31.7\\
						supnorm & 0.924(0.025) & 0.52 & 32.6 & 36.6 & 36.1 & 36.4 & 36.2& 0.848(0.020) & 13.93 & 46.6 & 34.2 & 33.8 & 33.7 & 33.5\\\hline
						Models&\multicolumn{14}{|c|}{Correlation $\rho=0.8$}\\\hline
						elastic net & 0.801(0.018) & 0.19 & 24.1 & 29.6 & 29.7 & 30.6 & 29.6& 0.773(0.036) & 3.74 & 15.7 & 35.4 & 36.3 & 36.0 & 35.7\\
						group Lasso & 0.761(0.023) & 0.38 & 64.0 & 21.4 & 21.2 & 21.3 & 21.2 & 0.654(0.023) & 12.30 & 89.7 & 17.3 & 17.6 & 17.5 & 17.3\\
						supnorm & 0.743(0.023) & 0.45 & 63.1 & 34.1 & 34.0 & 33.9 & 34.2& 0.667(0.016) & 14.01 & 79.8 & 35.3 & 35.3 & 35.3 & 35.2\\\hline
					\end{tabular}}
				\end{center}}\vspace{-0.3cm}
			\end{table}							
									
									The second test is a four-class example with each sample in $p$-dimensional space. The data in class $j$ was generated from the mixture Gaussian distribution $\mcn(\bm{\mu}_j,\bm{\Sigma}_j), j = 1,2,3,4$. The mean vectors and covariance matrices are $\bm{\mu}_2=-\bm{\mu}_1, \bm{\mu}_4=-\bm{\mu}_3, \bm{\Sigma}_2=\bm{\Sigma}_1, \bm{\Sigma}_4=\bm{\Sigma}_3$, and
									{\vspace{-0.2cm}\small \begin{align*}\bm{\mu}_1&=(\underset{s}{\underbrace{1,\cdots,1}},\underset{p-s}{\underbrace{0,\cdots,0}})^\top,\
										\bm{\mu}_3=(\underset{s/2}{\underbrace{0,\cdots,0}},\underset{s}{\underbrace{1,\cdots,1}},\underset{p-3s/2}{\underbrace{0,\cdots,0}})^\top,\\
										\bm{\Sigma}_1&=\begin{bmatrix}
										\rho\mathbf{E}_{s\times s}+(1-\rho)\mbfi_{s\times s} &\\
										& \mbfi_{(p-s)\times (p-s)}\end{bmatrix},\\ \bm{\Sigma}_3&=\begin{bmatrix}
										\mbfi_{\frac{s}{2}\times \frac{s}{2}} & &\\
										& \rho\mathbf{E}_{s\times s}+(1-\rho)\mbfi_{s\times s} &\\
										& & \mbfi_{(p-\frac{3s}{2})\times (p-\frac{3s}{2})}
										\end{bmatrix}.\end{align*}}This kind of data was also tested in \cite{wang2008hybrid,ye2011efficient} for binary classifications. We took $p=500, s=30$ and $\rho=0, 0.8$ in this test. As did in last test, the best parameters for all models were tuned by first generating $n=100$ training samples and another $n=100$ validation samples. Then we compared the different models solved by ADMM and Sedumi with the selected parameters on $n=100$ randomly generated training samples and $n'=20,000$ random testing samples. The comparison was independently repeated 100 times. The performance of different models and algorithms were measured by prediction accuracy, running time (sec), the number of incorrect zeros (IZ), the number of nonzeros in each column (NZ1, NZ2, NZ3, NZ4), where IZ, NZ1, NZ2, NZ3, NZ4 were counted in a similar way as that in last test by first truncating the output solution $\mbfw$. Table \ref{table:syn2} lists the average results, from which we can see that the elastic net MSVM tends to give best predictions. ADMM is much faster than Sedumi, and interestingly, ADMM also gives higher prediction accuracies than those by Sedumi. This is probably because the solutions given by Sedumi are sparser and have more IZs than those by ADMM.  
									
	\begin{table}\caption{Original distributions of SRBCT and leukemia data sets}\label{table:distribution}\vspace{-0.5cm}
				{\small
					\begin{center}
						\scalebox{0.8}{\begin{tabular}{|c|ccccc||cccc|}\hline
								\multirow{2}{*}{Data set}&\multicolumn{5}{|c||}{SRBCT}&\multicolumn{4}{|c|}{leukemia}\\\cline{2-10}
								& NB & RMS &BL & EWS &total & B-ALL & T-ALL & AML & total\\\hline
								Training & 12 & 20 & 8 & 23 & 63 & 19 & 8 & 11 & 38\\
								Testing & 6 & 5 & 3 & 6 & 20 & 19 & 1 & 14 & 34\\\hline
							\end{tabular}}
						\end{center}}\vspace{-1.0cm}
					\end{table}

%
									\subsection{Real data}
									This subsection tests the three different MSVMs on microarray classifications. Two real data sets were used. One is the children cancer data set in \cite{khan2001classification}, which used cDNA gene expression profiles and classified the small round blue cell tumors (SRBCTs) of childhood into four classes: neuroblastoma (NB), rhabdomyosarcoma (RMS), Burkitt lymphomas (BL) and the Ewing family of tumors (EWS).  The other is the leukemia data set in \cite{golub1999molecular}, which used gene expression monitoring and classified the acute leukemias into three classes: B-cell acute lymphoblastic leukemia (B-ALL), T-cell acute lymphoblastic leukemia (T-ALL) and acute myeloid leukemia (AML). The original distributions of the two data sets are given in Table \ref{table:distribution}. Both the two data sets have been tested before on certain MSVMs for gene selection; see \cite{wang20071,zhang2008variable} for example.

									Each observation in the SRBCT dataset has dimension of $p=2308$, namely, there are 2308 gene profiles. We first standardized the original training data in the following way. Let $\mbfx^o=[\bfx_1^o,\cdots,\bfx_n^o]$ be the original data matrix. The standardized matrix $\mbfx$ was obtained by
									{\small $$\vspace{-0.1cm}x_{gj}=\frac{x_{gj}^o-\text{mean}(x_{g1}^o,\cdots,x_{gn}^o)}{\text{std}(x_{g1}^o,\cdots,x_{gn}^o)},\ \forall g, j.$$}Similar normalization was done to the original testing data. Then we selected the best parameters of each model by three-fold cross validation on the standardized training data. The search range of the parameters is the same as that in the synthetic data tests. Finally, we put the standardized training and testing data sets together and randomly picked 63 observations for training and the remaining 20 ones for testing. The average prediction accuracy, running time (sec), number of non-zeros (NZ) and number of nonzero rows (NR) of 100 independent trials are reported in Table \ref{table:SRBCT}, from which we can see that all models give similar prediction accuracies. ADMM produced similar accuracies as those by Sedumi within less time while Sedumi tends to give sparser solutions because Sedumi is a second-order method and more accurately solves the problems.
									
	\begin{table}\caption{Results of different models solved by ADMM and Sedumi on SRBCT and Leukemia data sets}
						\label{table:SRBCT}
						{\small
							\begin{center}
								\scalebox{0.8}{\begin{tabular}{|c|c|cccc||cccc|}\hline
										\multirow{2}{*}{Data}&\multirow{2}{*}{Models}& \multicolumn{4}{|c||}{ADMM} & \multicolumn{4}{|c|}{Sedumi}\\\cline{3-10}
										& &Accuracy & time & NZ & NR & Accuracy & time & NZ & NR\\\hline
										\multirow{3}{*}{SRBCT} & elastic net & 0.996(0.014) & 1.738 & 305.71 & 135.31 & 0.989(0.022) & 8.886 & 213.67 & 96.71 \\
										& group Lasso & 0.995(0.016) & 2.116 & 524.88 & 137.31 & 0.985(0.028) & 42.241 & 373.44 & 96.27 \\
										& supnorm & 0.996(0.014) & 3.269 & 381.47 & 114.27 & 0.990(0.021) & 88.468 & 265.06 & 80.82 \\\hline\hline
										
										\multirow{2}{*}{Leukemia} & elastic net & 0.908(0.041) & 1.029 & 571.56 & 271.85 & 0.879(0.048) & 30.131 & 612.16 & 291.71 \\
												& group Lasso & 0.908(0.045) & 2.002 & 393.20 & 150.61 & 0.838(0.072) & 76.272 & 99.25 & 44.14\\
												& supnorm & 0.907(0.048) & 2.211 & 155.93 & 74.60 & 0.848(0.069) & 121.893 & 86.03 & 41.78 \\\hline
									\end{tabular}}
								\end{center}}\vspace{-0.3cm}
							\end{table}								
									
									The leukemia data set has $p=7,129$ gene profiles. We standardized the original training and testing data in the same way as that in last test. Then we rank all genes on the standardized training data by the method used in \cite{dudoit2002comparison}. Specifically, let $\mbfx=[\bfx_1,\cdots,\bfx_n]$ be the standardized data matrix. The relevance measure for gene $g$ is defined as follows:
									{\small $$\vspace{-0.2cm}R(g)=\frac{\sum_{i,j}I(y_i=j)(m_{g}^j-m_g)}{\sum_{i,j}I(y_i=j)(x_{gi}-m_{g}^j)},\
									g = 1,\cdots,p,$$}where $m_g$ denotes the mean of $\{x_{g1},\cdots,x_{gn}\}$ and $m_{g}^j$ denotes the mean of $\{x_{gi}: y_i=j\}$. According to $R(g)$, we selected the 3,571 most significant genes. Finally, we put the processed training and tesing data together and randomly chose 38 samples for training and the remaining ones for testing. The process was independently repeated 100 times. Table \ref{table:SRBCT} tabulates the average results, which show that all three models give similar prediction accuracies. ADMM gave better prediction accuracies than those given by Sedumi within far less time. The relatively lower accuracies given by Sedumi may be because it selected too few genes to explain the diseases.

\section{Conclusion}\label{sec:conclusion}
We have developed an efficient unified algorithmic framework for using ADMM to solve regularized MSVS. By effectively using the Woodbury matrix identity we have substantially reduced the computational effort required to solve large-scale MSVMS. 
Numerical experiments on both synthetic and real data demonstrate the efficiency of ADMM by comparing it with the second-order method Sedumi. 


%
%


\begin{thebibliography}{10}
	
	\bibitem{BishopPRMLbook2006}
	Bishop C.:
	\newblock Pattern Recognition and Machine Learning,
	\newblock Springer-Verlag, New York (2006).
	
	\bibitem{bottou1994comparison}
	Bottou~L., Cortes~C., Denker~J.S., Drucker~H., Guyon~I., Jackel~L.D., LeCun~Y.,
	Muller U.A., Sackinger E., Simard P., et~al.:
	\newblock Comparison of classifier methods: a case study in handwritten digit
	recognition.
	\newblock In { Proceedings of the 12th IAPR International Conference on
		Pattern Recognition}, volume~2, pages 77--82 (1994).
	
	\bibitem{boydADMMbook}
	Boyd~S., Parikh~N., Chu~E., Peleato~B., and Eckstein~J.:
	\newblock {Distributed Optimization and Statistical Learning via the Alternating Direction Method of Multipliers}.
	\newblock {Foundations and Trends in Machine Learning}, 3(1):1--122 (2010).
	
	
	\bibitem{bradley1998feature}
	Bradley P.S. and Mangasarian O.L.:
	\newblock Feature selection via concave minimization and support vector
	machines.
	\newblock In { Proceedings of the Fifteenth International Conference of
		Machine Learning (ICML'98)}, pages 82--90 (1998).
	
	\bibitem{chen2009accelerated}
	Chen X., Pan W., Kwok J.T., and Carbonell J.G.:
	\newblock Accelerated gradient method for multi-task sparse learning problem.
	\newblock In { Proceedings of the Ninth International Conference on Data
		Mining (ICDM'09)}, pages 746--751. IEEE (2009).
	
	\bibitem{cortes1995support}
	Cortes C. and Vapnik V.:
	\newblock Support-vector networks.
	\newblock { Machine learning}, 20(3):273--297 (1995).
	
	\bibitem{crammer2002algorithmic}
	Crammer K. and Singer Y.
	\newblock On the algorithmic implementation of multiclass kernel-based vector machines.
	\newblock {Journal of Machine Learning Research}, 2: 265--292 (2002).

	
	\bibitem{deng2012}
	Deng W. and Yin W.:
	\newblock On the global and linear convergence of the generalized alternating
	direction method of multipliers.
	\newblock { Rice technical report TR12-14} (2012).
	
	\bibitem{dudoit2002comparison}
	Dudoit S., Fridlyand J., and Speed T.P.:
	\newblock Comparison of discrimination methods for the classification of tumors
	using gene expression data.
	\newblock { Journal of the American statistical association},
	97(457):77--87 (2002).
	
%
	\bibitem{glowinski2008numerical}
	Glowinski R.:
	\newblock { Numerical methods for nonlinear variational problems}.
	\newblock Springer Verlag (2008).
	
	\bibitem{golub1999molecular}
	Golub~T.R., Slonim~D.K., Tamayo~P., Huard~C., Gaasenbeek~M., Mesirov~J.P.,
	Coller~H., Loh~M.L., Downing~J.R., Caligiuri~M.A., Bloomfield~C.D., and Lander~E.S.:
	\newblock Molecular classification of cancer: class discovery and class
	prediction by gene expression monitoring.
	\newblock { Science}, 286(5439):531--537 (1999).
	
	\bibitem{grant2008cvx}
	Grant M. and Boyd S.:
	\newblock {CVX} - {Matlab} software for disciplined convex programming, version
	2.1.
	\newblock { http://cvxr.com/cvx} (2014).
	
	\bibitem{hager1989updating}
	Hager W.W.:
	\newblock Updating the inverse of a matrix.
	\newblock { SIAM Review}, 31:221--239 (1989).
	
	
	\bibitem{hsu2002comparison}
	Hsu C.W. and Lin C.J.:
	\newblock A comparison of methods for multiclass support vector machines.
	\newblock { Neural Networks, IEEE Transactions on}, 13(2):415--425 (2002).
	
	
	\bibitem{khan2001classification}
	Khan J., Wei J.S., Ringn{\'e}r M., Saal L.H., Ladanyi M., Westermann F.,
	Berthold F., Schwab M., Antonescu C.R., Peterson C., et~al.:
	\newblock Classification and diagnostic prediction of cancers using gene
	expression profiling and artificial neural networks.
	\newblock { Nature medicine}, 7(6):673--679 (2001).
	
%
	\bibitem{lee2004multicategory}
	Lee Y., Y.~Lin, and Wahba G.:
	\newblock Multicategory support vector machines.
	\newblock { Journal of the American Statistical Association},
	99(465):67--81 (2004).
	
	
	
	\bibitem{platt2000large}
	Platt J.C., Cristianini N., and Shawe-Taylor J.:
	\newblock Large margin dags for multiclass classification.
	\newblock { Advances in neural information processing systems},
	12(3):547--553 (2000).
	
	
	%
	\bibitem{sturm1999using}
	Sturm J.:
	\newblock Using {SeDuMi} 1.02, a {MATLAB} toolbox for optimization over
	symmetric cones.
	\newblock { Optimization methods and software}, 11(1-4):625--653 (1999).
	
	\bibitem{wang20071}
	Wang L. and Shen, X.:
	\newblock On ${L}_1$-norm multiclass support vector machines.
	\newblock { Journal of the American Statistical Association},
	102(478):583--594 (2007).
	
	\bibitem{wang2008hybrid}
	Wang L., Zhu J., and Zou, H.:
	\newblock Hybrid huberized support vector machines for microarray
	classification and gene selection.
	\newblock { Bioinformatics}, 24(3):412--419 (2008).
	
	
	
	\bibitem{ye2011efficient}
	Ye G.B., Chen Y., and Xie X.:
	\newblock Efficient variable selection in support vector machines via the
	alternating direction method of multipliers.
	\newblock In { Proceedings of the International Conference on Artificial
		Intelligence and Statistics} (2011).
	
	\bibitem{yuan2006model}
	Yuan M. and Lin, Y.:
	\newblock Model selection and estimation in regression with grouped variables.
	\newblock { Journal of the Royal Statistical Society: Series B (Statistical
		Methodology)}, 68(1):49--67 (2006).
	
	\bibitem{zhang2008variable}
	Zhang H., Liu Y., Wu Y., and Zhu J.:
	\newblock Variable selection for the multicategory svm via adaptive sup-norm
	regularization.
	\newblock { Electronic Journal of Statistics}, 2:149--167 (2008).
	
\end{thebibliography}

\end{document}